\documentclass{article}

\usepackage{microtype}
\usepackage{graphicx}
\usepackage{subfigure}
\usepackage{booktabs} 
\usepackage{xspace}
\usepackage{amsmath,amsfonts,amssymb,amsthm}
\usepackage[dvipsnames]{xcolor}
\usepackage{diagbox}
\usepackage{pifont}
\usepackage{amsmath}
\usepackage{natbib}
\usepackage{enumitem}
\usepackage{times}
\usepackage{epsfig}

\usepackage[pagebackref=true,breaklinks=true,colorlinks,bookmarks=false]{hyperref}

\usepackage{hyperref}

\usepackage[accepted]{icml2020}

\icmltitlerunning{Spectral Roll-off Points Variations: Exploring Useful Information in Feature Maps by Its Variations}

\begin{document}
\setlength{\textfloatsep}{10pt plus 1.0pt minus 2.0pt}

\twocolumn[

\icmltitle{Spectral Roll-off Points Variations: Exploring Useful Information in Feature Maps by Its Variations}



\icmlsetsymbol{equal}{*}

\begin{icmlauthorlist}
\icmlauthor{Yunkai Yu}{bit_auto}
\icmlauthor{Yuyang You\textsuperscript{*}}{bit_auto}
\icmlauthor{Zhihong Yang}{implad}
\icmlauthor{Guozheng Liu}{bit_auto}
\icmlauthor{Peiyao Li}{thu,ghddi}
\icmlauthor{Zhicheng Yang}{pa}
\icmlauthor{Wenjing Shan}{bit_auto}
\end{icmlauthorlist}

\icmlaffiliation{bit_auto}{Beijing Institute of Technology, Beijing, China}
\icmlaffiliation{implad}{Institute of Medicinal Plant Development, Chinese Academy of Medical Sciences, Beijing, China}
\icmlaffiliation{thu}{Tsinghua University, Beijing, China}
\icmlaffiliation{ghddi}{Global Health Drug Discovery Institute, Beijing, China}
\icmlaffiliation{pa}{PAII Inc, Palo Alto, USA}

\icmlcorrespondingauthor{Yuyang You}{arthurwy@163.com}

\icmlkeywords{Spectral roll-off points,
Useful information, Frequency, Sampling theorem,
Convolutional Neural Network, Computer vision}

\vskip 0.3in
]



\printAffiliationsAndNotice



\makeatletter
\DeclareRobustCommand\onedot{\futurelet\@let@token\@onedot}
\def\@onedot{\ifx\@let@token.\else.\null\fi\xspace}

\def\eg{\emph{e.g}\onedot} \def\Eg{\emph{E.g}\onedot}
\def\ie{\emph{i.e}\onedot} \def\Ie{\emph{I.e}\onedot}
\def\cf{\emph{c.f}\onedot} \def\Cf{\emph{C.f}\onedot}
\def\etc{\emph{etc}\onedot} \def\vs{\emph{vs}\onedot}
\def\wrt{w.r.t\onedot} \def\dof{d.o.f\onedot}
\def\etal{\emph{et al}\onedot}
\def\viz{\emph{viz}\onedot}
\makeatother



\newcommand{\cmark}{\ding{51}}%
\newcommand{\xmark}{\ding{55}}%


\begin{abstract}
Useful information (UI) is an elusive concept in neural networks. A quantitative measurement of UI is absent, despite the variations of UI can be recognized by prior knowledge. The communication bandwidth of feature maps decreases after downscaling operations, but UI flows smoothly after training due to lower Nyquist frequency. Inspired by the low-Nyqusit-frequency nature of UI, we propose the use of spectral roll-off points (SROPs) to estimate UI on variations. The computation of an SROP is extended from a 1-D signal to a 2-D image by the required rotation invariance in image classification tasks. SROP statistics across feature maps are implemented as layer-wise useful information estimates. We design sanity checks to explore SROP variations when UI variations are produced by variations in model input, model architecture and training stages. The variations of SROP is synchronizes with UI variations in various randomized and sufficiently trained model structures. Therefore, SROP variations is an accurate and convenient sign of UI variations, which promotes the explainability of data representations with respect to frequency-domain knowledge.
\end{abstract}

With prior knowledge, \textbf{variations of useful information} can be clear, despite the amount of useful information remains opaque. Downsampling blocks is a common block in DNNs, it filters high-frequencies directly. The sampling theorem states that these operations, which always halve the spatial resolution (sampling frequency), lead to a decline of Nyquist frequency in consecutive feature maps \cite{cover,shannon}. Well-trained models possess more useful information than randomized models, so their downsampling blocks eliminate less useful information. Therefore, we can conclude that useful information in feature maps is in a low-Nyquist-frequency form. This property is denoted as low-Nyquist-frequency prior (LFP). Motivated by LFP, we propose the use of spectral roll-off points (SROPs), which quantify feature energy in low-frequency bands. It can be used as an estimate of useful information when its variations can be regarded as a sign of useful information variations. 

Some obstacles in computation need to be tackled in order to obtain estimates of 3-D feature maps, because SROP is computed by analyzing a 1-D spectrum of a 1-D signal. Transforming the 2-D spectrum to 1-D spectrum via a radial average method enables us to extend the SROP computation to a 2-D feature map. This modification is consistent with the desired property in image classification tasks that well-trained DNNs possess rotation invariance \cite{rt_invar_3,rt_invar_2,rt_invar_1}. SROP statistics among kernels are adopted in the estimation of layer-wise useful information. This design substantially simplifies the computation load.

Experiments provide empirical evidence which confirms that SROPs variations can trace the variations of useful information. Three factors that relate to useful information variations, model input, model structures and sufficient training, are summarized from previous work \cite{rt_invar_2,Understanding,why_useful,Yak2019TowardsTA}. Control experiments are designed under the sanity check framework \cite{Sanity_Checks_for_Saliency_Maps}. Everyday-object and digit images are synthesized with different proportions, which produces variations in the patterns of useful information or in noise intensity. The SROPs of feature maps see a synchronous change. The use of SROP statistics explains the effectiveness of downscaling, batch normalization (BN), anti-aliased blocks and intermediate layers. Layer-wise SROP curves visualize the flows of useful information in multiple modern model architectures. Comparisons between randomized and pre-trained models demonstrate that low-Nyquist-frequency data representations are the results of sufficient training. Variations of SROPs and useful information are proved to be consistent. The feasibility of SROP statistics, potential applications and limitations of SROPs are discussed in Sec. \ref{sec:discussion}.

Our contributions are summarized as follows:
\begin{itemize}
    \item We measure useful information by its variations, and propose the use of SROPs variations according to LFP. The computation of SROPs is extended from a 1-D signal to DNN feature maps. Consequently, we can explain layer-wise useful information with frequency-domain knowledge.
    \item Systematic analysis proves that SROP variations is a sign of useful information variations in a layer-wise level. The cause of useful information variation covers model input, model structures and sufficient training. Experiments include 20 modern DNNs and 3 large-scale datasets.
    \item Theoretical analysis and empirical evidence show that SROP separates redundant information from useful information. The computation of SROP requires only a single sample, which makes it a practical layer-wise measurement to promote the explainability of DNNs.
\end{itemize}
\section{Related work}
\label{sec:related_work}
The most related studies are the ones that evaluate information, where useful information and redundancy coexist in DNN representations and are assessed together. We also demonstrate the essence of LFP from existing observations. LFP interprets information on frequency domain. Therefore, frequency analysis is included in this section. 
\subsection{Metrics of evaluating information} Current metrics based on information theory evaluate transmissible information. Shannon mutual information measures the entropy variation between two variables. Therefore, entropy including Renyi and $H$ entropy \cite{h_entropy,renyi}, associates the information to prediction loss. However, using these metrics in high dimension meets difficulties in computation. $\mathcal{V}$-predictive information \cite{A_Theory_of_Usable_Information_under_Computational_Constraints} concentrates on actionable information. It calibrates mutual information through an additional benchmark entropy from natural random process and extends the application scope. Data-dependent priors are needed to improve the use of mutual information \cite{NEURIPS2019_05ae14d7}. Otherwise, the outputs of different DNN layers contain spurious causal relationships. It indicates that computing mutual information with variables in different layers may capture spurious causal relationships.

Assessing the representational capacity of a hidden layer is another method to evaluate existing information in DNN feature maps. Intrinsic dimensionality (ID) is the minimal number of base vectors to reconstruct hidden manifold. Projection methods search a best subspace based on the projection error or local connectivity to compute ID \cite{pca,local_conn}. Nearest-neighbour ID estimators infer the fractal dimension by first neighbors under the assumptions of data neighbourhood distribution \cite{frac}. A recent TWO-NN estimator shows a satisfactory universality \cite{nature_ID}. As a statistical method, the accuracy of ID estimation is inevitably compromised by the scale and heterogeneity of datasets. Jiang \textit{et al.} has reviewed a wide rang of complexity measures, and most of them are negatively correlated with generalization, an important property that we expect in useful information \cite{JiangNMKB20}.

Ignoring LFP is the foremost problem of mutual information and ID when estimating useful information. The consequence is that noise and redundancy, which are common in DNNs \cite{lottery}, are counted as useful components. On the basis of LFP, useful information is recognized as low-Nyquist-frequency representations, which inspires the use of SROPs.

\subsection{Importance of LFP}
The decrease in spatial resolution of feature maps guarantees the correctness of LFP. LFP implies that useful information is always in the form of low-frequency feature maps if the model is well-trained. Therefore, low-frequency data representations are more likely to be useful. An evidence is that anti-aliased models implement additional low-pass filters in downscaling blocks but received unexpected improved model performance \cite{make_conv_shift_inva}. Unlike useful information, feature maps of noise can have high frequency components that will be discarded by e.g. a following max-pooling block. The redundancy in high-frequency band motivates a sparse sampling method in feature maps to reduce computation \cite{spatially_adaptive}. Kernel smoothness is highlighted in encoding high-frequency patterns of input in adversarial learning \cite{A_Theory_of_Usable_Information_under_Computational_Constraints}.  Wang \etal encourage using smooth kernels to make models generalizable~\cite{High-Frequency-Component-Helps-Explain-the}.
These works address the importance of low-frequency data representation in the training process and thereby gain sufficient useful information (excellent model performance). Hence, LFP is an important and generic property (LFP) of DNNs. 

\subsection{Frequency analysis} 
Frequency analysis has a more direct relation with spatial resolution compared to low-dimension prior that probability mass concentrates near regions with a much smaller dimensionality than the original space where the data live \cite{Representation_Learning}. Current frequency analysis has focused on the frequency components in model input. It's found that DNNs firstly learn low-frequency components of model input during training process \cite{freq_principle,training_behavior}. The \textit{spectral bias} indicates that high-frequency functions of input can be expressed by low frequency network functions defined in hidden manifolds \cite{on_the_spectral_bias}. Optimisation procedure is content-aware, where learning simple patterns of the data is the priority \cite{Arpit2017ACL}. By comparison, our frequency analysis is derived on the basis of LFP and aims at evaluating the useful information in DNN feature maps in terms of frequency domain.

\section{Preliminaries}
\paragraph{Ground truth of useful information}\label{sec:truth_ui} The ground truth of useful information is given by its required properties such as possessing visual concepts and \textit{knowledge consistency} \cite{knowledge_consistency_2,knowledge_consistency_1}. According to human ingenuity, model performance can be used as the ground truth of useful information in well-trained DNNs, because these models are able to assist humans in real-world applications. Useful information is content-aware. Therefore, modifying model input produces a useful information variation in DNN feature maps. Another property is that useful information along depth is non-increasing ~\cite{Opening_the_Black_Box_of_Deep_Neural_Networks_via_Information}. It can be used to discover redundancy (Sec. \ref{subsec:compar}). 
\paragraph{Notations of DNN} Throughout the paper a neural network backbone $F$ is denoted as a consequence of $L$ hidden layers $\{T_1, T_2, \cdots, T_L\}$ as follows: 
\begin{equation}
F(X)=(T_L\circ T_{L-1}\circ \cdots T_1)(X),   
\end{equation}
where $X$ is the model input. The output of the $k_{th}$ layer $Y_k$ is defined as $Y_k=(T_k\circ T_{k-1}\circ \cdots T_1)(X)$. $Y_k$ is a $\mathit{c}_k \times \mathit{n}_k\times \mathit{n}_k$ vector, composed of $\mathit{c}_k$ 2-D feature maps.
\paragraph{Notations of layers}The notations of layers are shown in Table \ref{tab:layer_name}. The kernel sizes of the pooling layers are two except for those of AlexNet and DenseNet, where the kernel size is three. The convolutional blocks in AlexNet and VGG16 contain a convolution and an activation function ReLU, while other convolutional layers have an additional BN layer. The pooling layers and stride-convolutions in anti-aliased models behave differently (see Sec. \ref{subsec:anti_ali}).
\paragraph{Anti-aliased blocks} \label{subsec:anti_ali}
Anti-aliased blocks in this study improve downscaling operations to make DNNs shift-invariant. They use a dense max-pooling layer to obtain an intermediate signal. Afterwards, the intermediate signal is downscaled by a blur-pooling layer to get the final output. The blur pooling layer used in our paper is equivalent to the bi-linear downsampling method. The effectiveness of the design is validated in large-scale databases like CIFAR10 and ImageNet \cite{cifa_data,ImageNet}. More details may be found in \cite{make_conv_shift_inva}.

\begin{table*}[h]
    \begin{center}
    \begin{tabular}{lccccc}
    \hline
    Backbone&AlexNet&{VGG}&\multicolumn{2}{c}{ResNet} & {DenseNet}\\

    Output size&&VGG16, VGG16-bn&ResNet18&ResNet34&\\
    \hline
      $224 \times 224$ &- & -& - & -& - \\
      & - &conv0.0-conv0.1 & -& -& -\\
      $112 \times 112$ &- & \textbf{pool1} &- &- &- \\
            &- &conv1.0-conv1.1& \textbf{conv0}&\textbf{conv0} &\textbf{conv0}\\
      $56 \times 56$ & -& \textbf{pool2}& \textbf{pool1}& \textbf{pool1}& \textbf{pool1}  \\
       ($55 \times 55$)& \textbf{conv0}& conv2.0-2.2&resblk1.0-1.1& resblk1.0-1.2& denseblk1\\
      $28 \times 28$ & \textbf{pool1}& \textbf{pool3} & \textbf{resblk2.0} & \textbf{resblk2.0}
      &\textbf{transblk1} \\
      ($27 \times 27$)&conv1.0& conv3.0-3.2& resblk2.1&resblk2.1-2.3& denseblk2\\
      $14 \times 14$ &\textbf{pool2}&  \textbf{pool4} &\textbf{resblk3.0}&\textbf{resblk3.0}&\textbf{transblk2} \\
      ($13 \times 13$)&conv2.0-2.2& conv4.0-4.2 &resblk3.1 &resblk3.1-3.5& denseblk3\\
      $7 \times 7$ & \textbf{pool3}& \textbf{pool5}&\textbf{resblk4.0} &\textbf{resblk4.0}&\textbf{transblk3}  \\
      & - & -& resblk4.1& resblk4.1-4.2& denseblk4, bn1\\ 
      \hline
    \end{tabular}
    \end{center}
     \caption{Notations of backbones. Notably, the \textit{convx.y} block in VGG16-bn has an additional BN layer compared to VGG16. The downscaled layers of original and anti-aliased models share the same notations and are highlighted in bold. The \textit{denseblk} and \textit{transblk} in DenseNet of different scales (DenseNet121, 169), also share the same notations.} \label{tab:layer_name}
\end{table*}
\section{Methods}
\label{methods}
SROPs characterize 1-D signals. The computation of SROPs is extrapolated to multiple 2-D feature maps generated from DNN kernels to estimate layer-wise useful information. Radially averaged power spectrum enables us to calculate a SROP in a 2-D feature map. DNN kernels outputs 3-D feature maps in a hidden layer. SROP statistics are used as the evaluation metrics of useful information, since they feature in the SROP distribution across kernels.
\subsection{SROPs of 1-D signal}
SROPs are known to be high for right-skewed distributions \cite{SROP} and low when data representations contain increased low-frequency components. Useful information is available only in low-frequency representations when followed by a downsampling block, whereas redundant information distributes randomly. Theoretically speaking, low SROPs indicate that useful information is ideally compressed. The SROP of a 1-D signal is defined in Eq. \ref{eq:srop}.
\begin{equation}
\label{eq:srop}
\sum_{k=b_1}^{SROP}S_{1D}(k)=\kappa \sum_{k=b_1}^{b_2}S_{1D}(k),
\end{equation}
where $S_{1D}(k)$ denotes the value of spectrum $S_{1D}$ in bin $k$, and $b_1$ and $b_2$ are the edges of the frequency band. $\kappa=0.85$ is a cut-off point of total energy. In this study, $S_{1D}$ is a one-sided power spectrum density, which is expressed as 
\begin{equation}
    S_{1D}(k)=\frac{|\mathcal{F}(k)|}{\sum^{b_2}_{j=b_1}{|\mathcal{F}(j)|}},
\end{equation}
where $\mathcal{F}(k)$ is the $k^{th}$ component of the Fourier sequence of input.
To make fair SROP comparisons among samples with different spatial resolutions, we normalized SROPs to $[0,1]$ as shown in Eq \ref{eq:srop_norm}: \begin{equation}\label{eq:srop_norm}SROP_{normalized}=(SROP-b_1)/(b_2-b_1).\end{equation}

\subsection{Radially averaged power spectrum}
We map the 2-D power spectrum to its radially averaged form to calculate the SROP of a 2-D image, which is based on the prior that well-trained models are useful for tasks require good rotational invariance. Actually, radially averaged SROP has been applied to estimate systematic errors in the noise levels of satellite images \cite{2dSROP}. A 2-D spectrum is denoted as $S_{2D}\in R^{n\times n}$ as $S_{2D}(r,\theta)$ in the polar basis, and the ordinate origin is at the center of the spectrum. After discretization, we obtain
\begin{equation}
    m=Floor(\frac{\sqrt{2}(n-1)}{2})-1,
\end{equation}
where $m$ is the number of radius bins of discretized 1-D spectrum, and $Floor(\cdot)$ takes the nearest integer value of input towards minus infinity. Subsequently, we establish a mapping relation to transform a 2-D spectrum $S_{2D}$ to 1-D spectrum $S_{1D}\in R^{1\times m}$, which can be formulated as
\begin{equation}
    S_{1D}(k)=\sum_{\theta=0}^{2\pi}S_{2D}(r_k,\theta), k\leq r_k<k+1, k\in [0,m],
\end{equation}
where $m\approx n\times 0.7$ when $n$ is large, suggesting a significant dip in the upper bound of frequency. This finding shows that the rotation invariance is a strong constraint.
When $X\rightarrow Rotate(X)$, we have
\begin{equation}
 S_{2D}(r,\theta)\rightarrow S_{2D}(r,\theta+\Delta \theta),
 S_{1D}(k)\rightarrow S_{1D}(k), \forall k.
\end{equation}
Therefore, rotating a 2-D image will rotate the 2-D spectrum $S_{2D}$, but the radially averaged 1-D spectrum $S_{1D}$ stays identical because it is independent with $\theta$.

\subsection{SROP statistics for layer-wise estimation}
\label{secsec:statistics}
DNN layers usually contain at least tens of filters. SROP distribution across kernels gives a comprehensive vision of layer-wise useful information. SROP statistics present the layer-wise useful information concisely. Kernel outputs of an ideal model are concentrated. Accordingly, a small internal covariate shift is encouraged \cite{bn}. It indicates that SROPs among kernels are close to their mean values for well-trained models. For the reason, SROP \textit{mean} value is used as the estimation of layer-wise useful information for the output $Y_k$. Median values and quartiles are provides to show that SROPs are concentrated. 

\begin{figure}
    \begin{center}
    \includegraphics[width=\columnwidth]{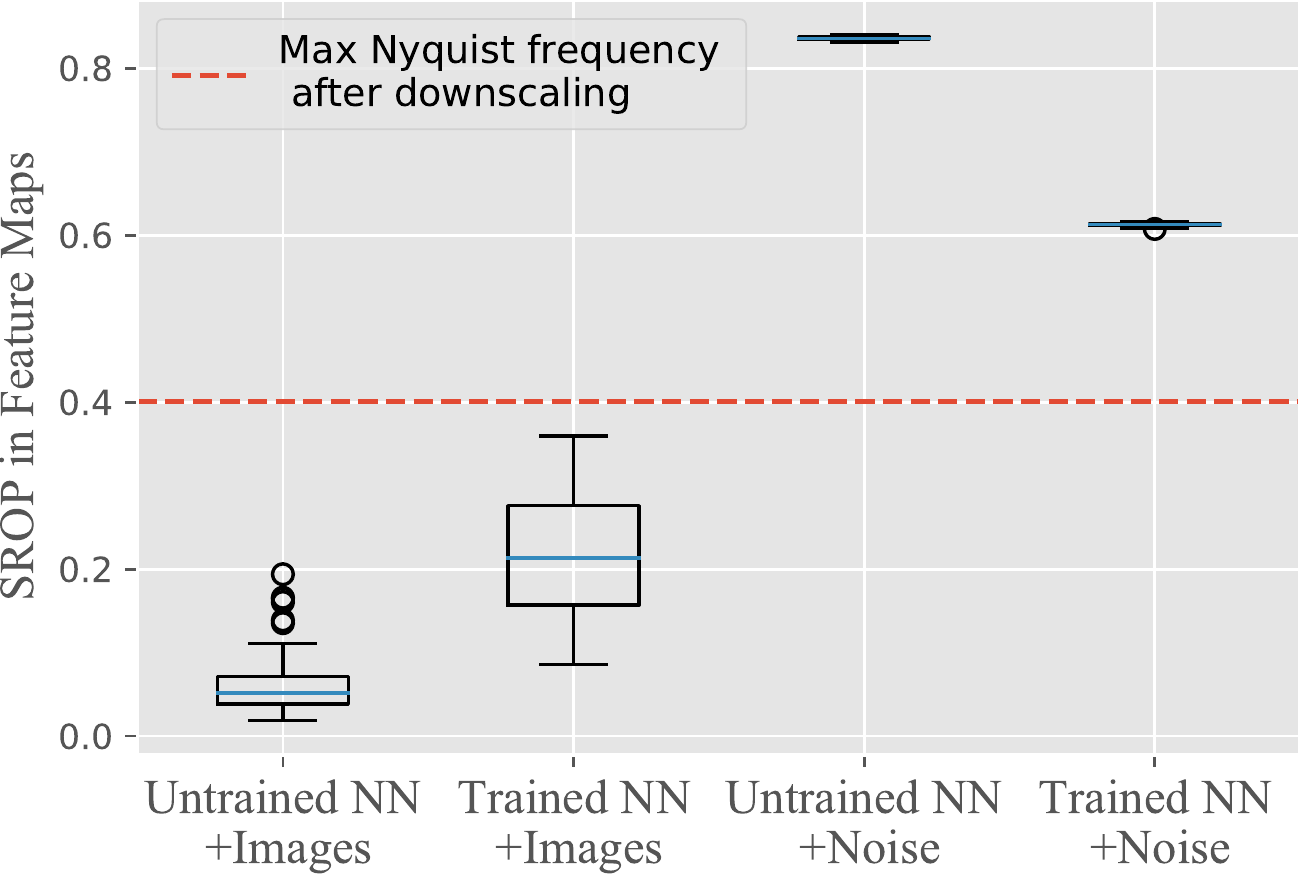}
    \end{center}
    \caption{A Demo of LFP. Images contain useful information, and they are low-Nyqusit-frequency. Noise, where useful information is mixed with plenty of redundant information are high-low-Nyquist-frequency. SROP is used as the Nyquist frequency of useful information. The SROP threshold reflects the highest Nyquist frequency of transmissible information. Information whose Nyquist frequency is higher than the threshold cannot pass to the consecutive layers perfectly. The demo shows that heterogeneity of samples is a confounding factor in obtaining accurate useful information, but the variations of useful information are trackable. According to LFP, a decrease in layer-wise SROP indicates an increase in useful information.}
    \label{fig:motivation}
\end{figure}

With SROP statistics, we are able to explain LFP, a property of useful information intuitively. As Figure \ref{fig:motivation} shows, the information of images are in a low-Nyquist form, so in all the intrinsic useful information can survive from the consecutive downscaling blocks considered to their low SROPs. As for noise, the redundant information is high-Nyquist-frequency. Feature maps in trained models contain more useful information than untrained ones. In feature maps with high SROPs, a decrease of SROP indicates a higher chance to survive for information. Therefore, we adopt \textit{a decrease in layer-wise SROP} as a sign of \textit{an increase in useful information}.

The use of SROP statistics is a simplification of kernel-wise SROP. It has risk of losing information of layer-wise SROP distribution and thereby may fail to differentiate DNN feature maps with different useful information. Thus, we analyze the feasibility of using SROP statistics with results from sufficient experiments in Sec. \ref{sec:discussion}.

\section{Experiment designs}
SROPs variations are designed to reflect the variations of useful information in data representations. Figure \ref{fig:motivation} provides a demo of LFP with SROP variations. Based on the ground truth of useful information (Sec. \ref{sec:truth_ui}), data and model randomized tests~\cite{Sanity_Checks_for_Saliency_Maps} are implemented to assess the correctness of SROPs on measuring useful information variations. To provide convincing useful information variations that are in accordance with model performance and human priority, we adopt classical tasks and publicly available models.
\subsection{Useful information variations from model input}
\label{subsec:ui_var_input}
\begin{figure*}
    \begin{center}
    \includegraphics[width=0.8\linewidth]{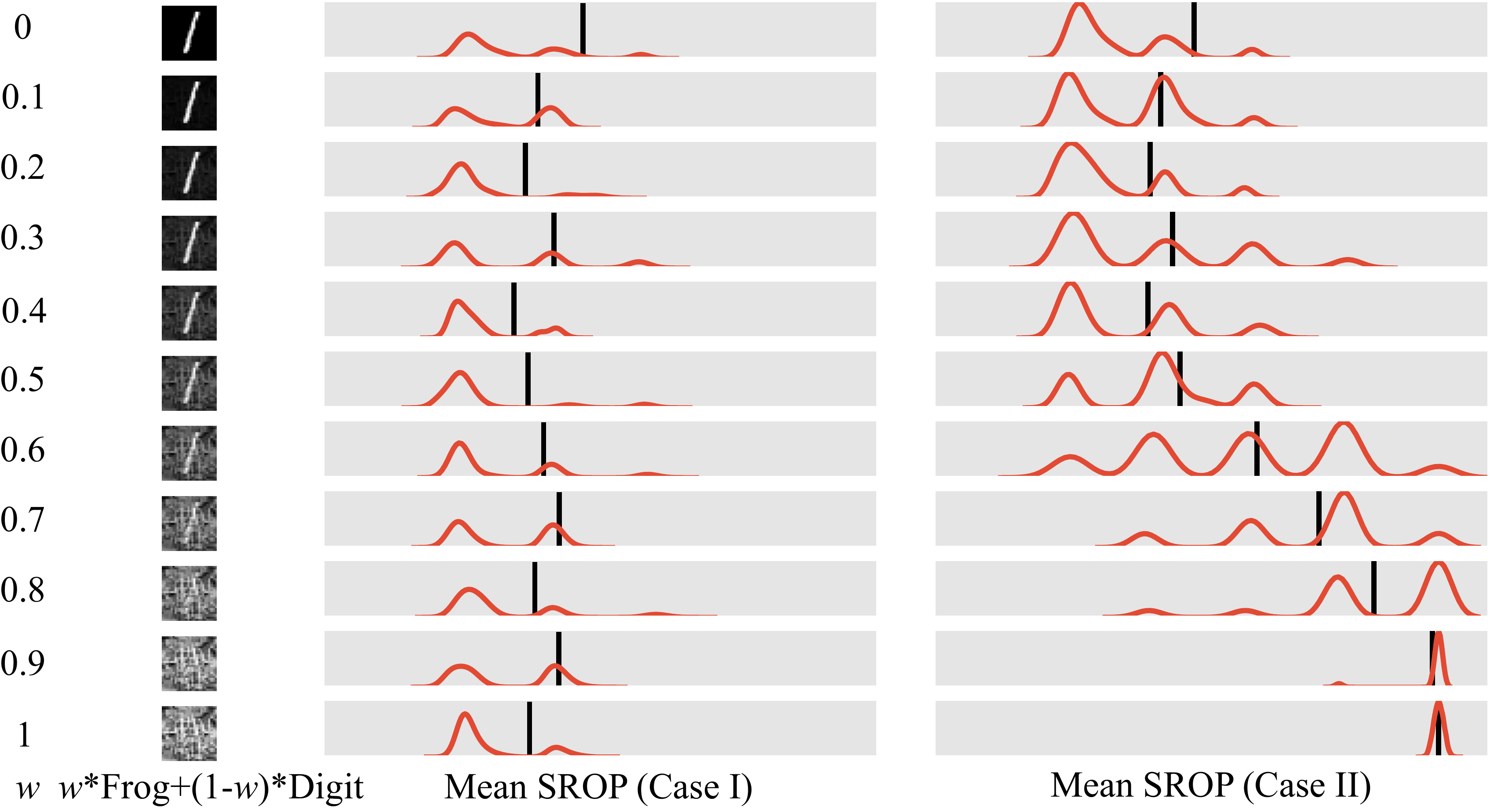}
    \end{center}
    \caption{Frog/digit images and SROP distributions. $w$ indicates the proportion of the frog image in synthesized images. The frog image is new useful information in CASE I and noise in CASE II when performing the digit classification task. Red plots are the mean SROPs' kernel density plots of the convolution when using the MNIST test set as model input. The models are well-trained. Vertical lines are mean SROPs when using the frog image as model input. Details of CASE I and II and the convolution can be found in Sec. \ref{subsec:ui_var_input}.}
    \label{fig:frog}
\end{figure*}

Variations of model input lead to variations of useful information in model representations. The variations of model input can be manifested by replacing previous useful information with different useful information or noise. Therefore, we design two cases of useful information variations as follows:
\begin{itemize}
    \item CASE I (Useful information variations from  different useful input): Embedding a frog image into samples of category "1", so $P(label=1|frog+digit)=1$;
    \item CASE II (Useful information variations from redundant information): Embedding a frog image into all digit images, so $P(label=1|frog+digit)=P(label=1|digit)$.
\end{itemize}
The feature maps of frog and digit images are called frog patterns and digit patterns, respectively. The modification of the model input, as shown in Figure \ref{fig:frog}, is the same in the two CASEs, but the variations of useful information in feature maps are from different resources. Frog patterns are useful information if they contribute to obtaining correct labels, otherwise they are noise. The frog patterns enables models to make correct decisions in CASE I, because they provide useful patterns as easy-to-learn hints to recognize samples in category "1". However, models are less likely to learn digit patterns when an image "1" is of lower proportion (higher $w$), which produces a variation of useful information. Therefore, CASE I exhibits the useful information variations from different useful model input. 

Meanwhile, the frog patterns are noise in CASE II, which erode useful digit patterns without contributing to model decision. In this way, the proportion of digit images $1-w$ reflects the degree of useful information variations in feature maps from redundant information. 

To ensure that the useful information variations conform to human prior, we adopt a classical CNN ~\cite{input_model}and simple datasets like MNIST and CIFAR10. The CNN is trained for 20 epochs with a learning rate of 0.001 in each CASEs. The SROP statistics of the last convolution of the well-trained model, and their variations in both CASEs are compared to the variations of useful information.

\subsection{Useful information variations from model structures and sufficient training}\label{secsec:max_baseline}
Downscaling layers lead to a sharp spatial resolution decline, and intermediate layers would modify data representations to preserve useful information. Layer-wise SROP statistics from various model structures are analyzed. A max-pooling block, where the stride and kernel size are both two, is used as the benchmark downscaling structure. Baseline SROPs are calculated from data representations when removing intermediate layers and then applying the benchmark structure. This process explores how intermediate layers facilitate the learning process by modifying data representations. The experiments include model structures like convolutions, max-pooling layers, blocks from ResNet and DenseNet, and anti-aliased blocks. Different model structures differentiate the useful information in data representations. Hence, the SROPs are supposed to behave differently.

Pre-trained models preserve more useful information than untrained models because they are sufficiently trained and useful in real-world applications. Therefore, we compare randomized and pre-trained models to analyze the effectiveness of sufficient training. In our experiments, models pre-trained on ImageNet are from PyTorch~\cite{pytorch} and Zhang~\cite{make_conv_shift_inva}. SROP curves visualize the SROP variation along depth, which makes it possible to explain the effect of model architectures on tuning useful information flows. The credibility of SROP will improve if it is compatible with observations about useful information. A known conclusion is that an increasing depth or width in a model would ease the extraction of useful information~\cite{on_the_spectral_bias}. Moreover, the accuracy gain from increased model complexity will diminish for large models~\cite{rt_invar_2,efficient_net}. We expect to implement an SROP analysis scheme to make these findings explainable.
\section{Results}
\label{sec:results}
\subsection{SROP variations from model input}\label{sec:results_input}
\begin{figure}[h]
    \begin{center}
    \includegraphics[width =\columnwidth]{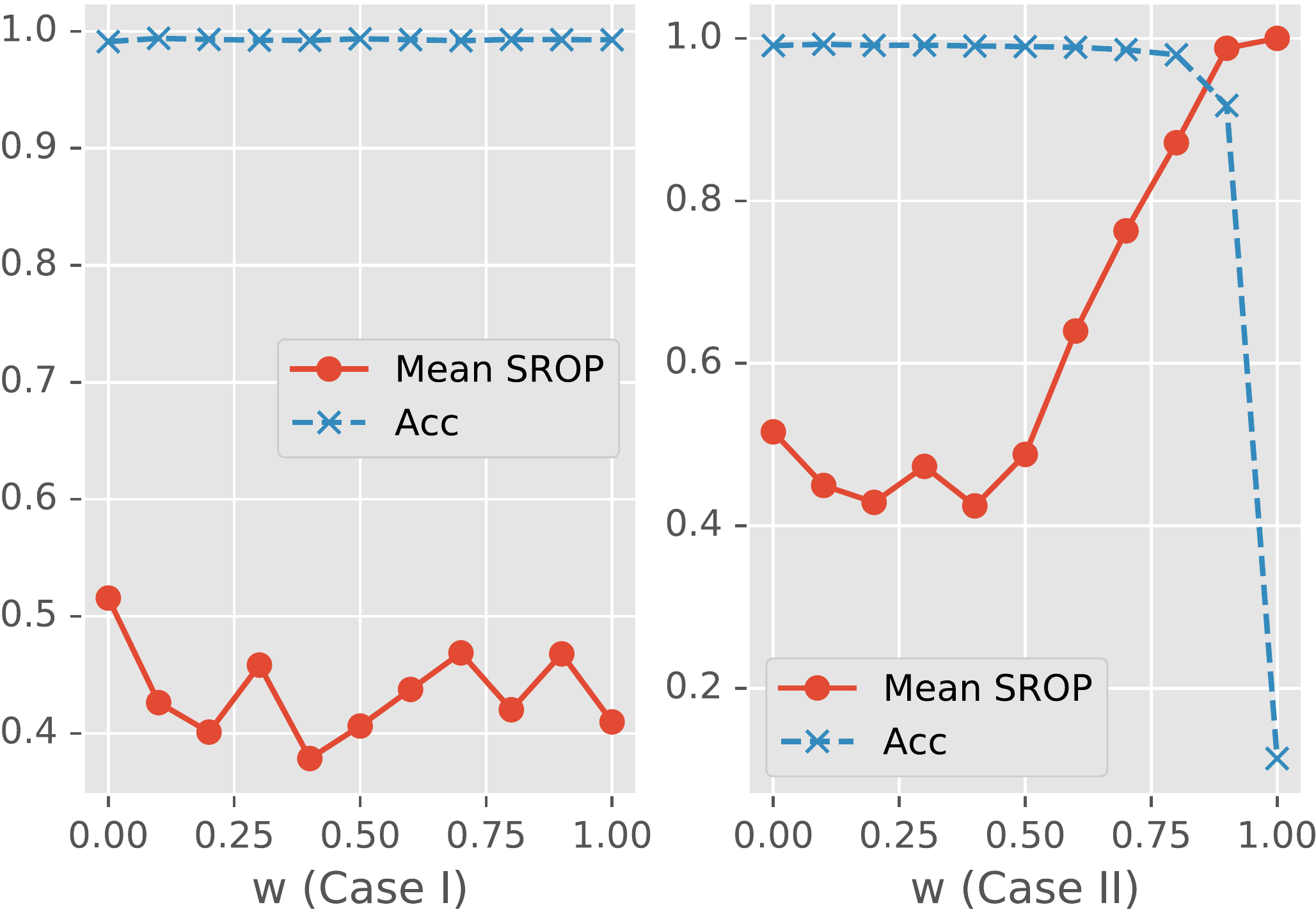}
    \end{center}
    \caption{Numerical values of accuracy and mean SROP of frog patterns. As introduced in Sec. \ref{subsec:ui_var_input}, the frog patterns are useful information and noise in CASE I and II respectively. The mean SROP in feature maps stays low when model extracts sufficient useful information for high accuracy (CASE I). The model fails to yield ideal accuracy (CASE II), which is reflected in high SROPs.}
    \label{fig:srop_acc}
\end{figure}

\paragraph{SROP variations from different useful patterns}
Digit and frog patterns are different useful patterns in CASE I. SROP distribution among samples is capable of reflecting the variation in model input. Figure \ref{fig:frog} demonstrates the kernel density (KDE) plot of SROPs of data representations among all the frog+digit samples. Digit patterns ($w=0$) has three peaks. Two low-SROP peaks is a characteristic of using frog patterns in model decision ($w=1$, CASE I). The SROP distributions of transition states ($0<w<1$, CASE I) have either two or three low-SROP peaks. In contrast, the SROPs of data representations rise to 1 when noise destroys useful digit patterns completely ($w=1$) in CASE II, as shown in the rightmost column. KDE plots with noisy input are likely to have additional peaks at high SROPs, as shown in CASE II. The variation of SROP distribution among samples is able to reflect the variation from different useful patterns. The phenomena support the idea that SROP can be used to estimate useful information in feature maps.
\paragraph{SROP variations from different noise intensity}
Figure \ref{fig:srop_acc} demonstrates that SROPs in feature maps are sensitive to noise intensity. The mean SROP of frog patterns increases monotonically as $w$ increase when $w>0.5$ in CASE II. Meanwhile, an apparent accuracy drop occurs when $w$ increases from 0.8 to 0.9. As a comparison, the same image generates low-SROP representations stably when it provides useful information (CASE I). SROPs are negatively related to noise intensity in model input.

\subsection{SROP variations from model structures and sufficient training}
\begin{figure}
    \begin{center}
    \includegraphics[width=\columnwidth]{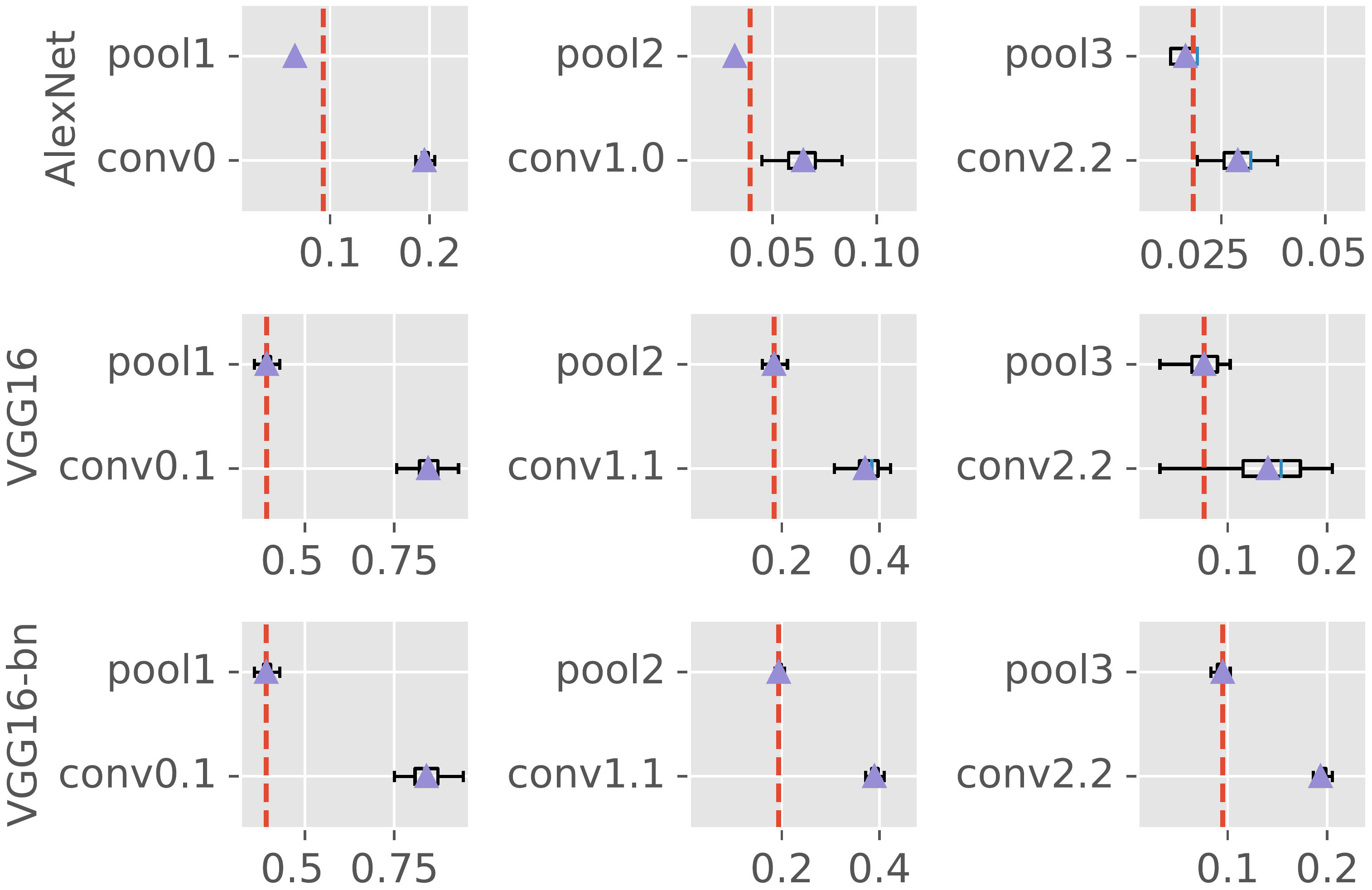}
    \end{center}
    \caption{Normalized SROPs in pooling layers and randomized strided-convolutions. The kernel sizes of the pooling layers in AlexNet and VGG are three and two, respectively. Triangles denote the SROP mean values. Dashed lines denote the SROP mean values from the benchmark (Sec. \ref{secsec:max_baseline}). Box plots contain median values and quartiles.}
    \label{fig:base}
\end{figure}

\begin{figure*}
    \begin{center}
    \includegraphics[width=0.8\linewidth]{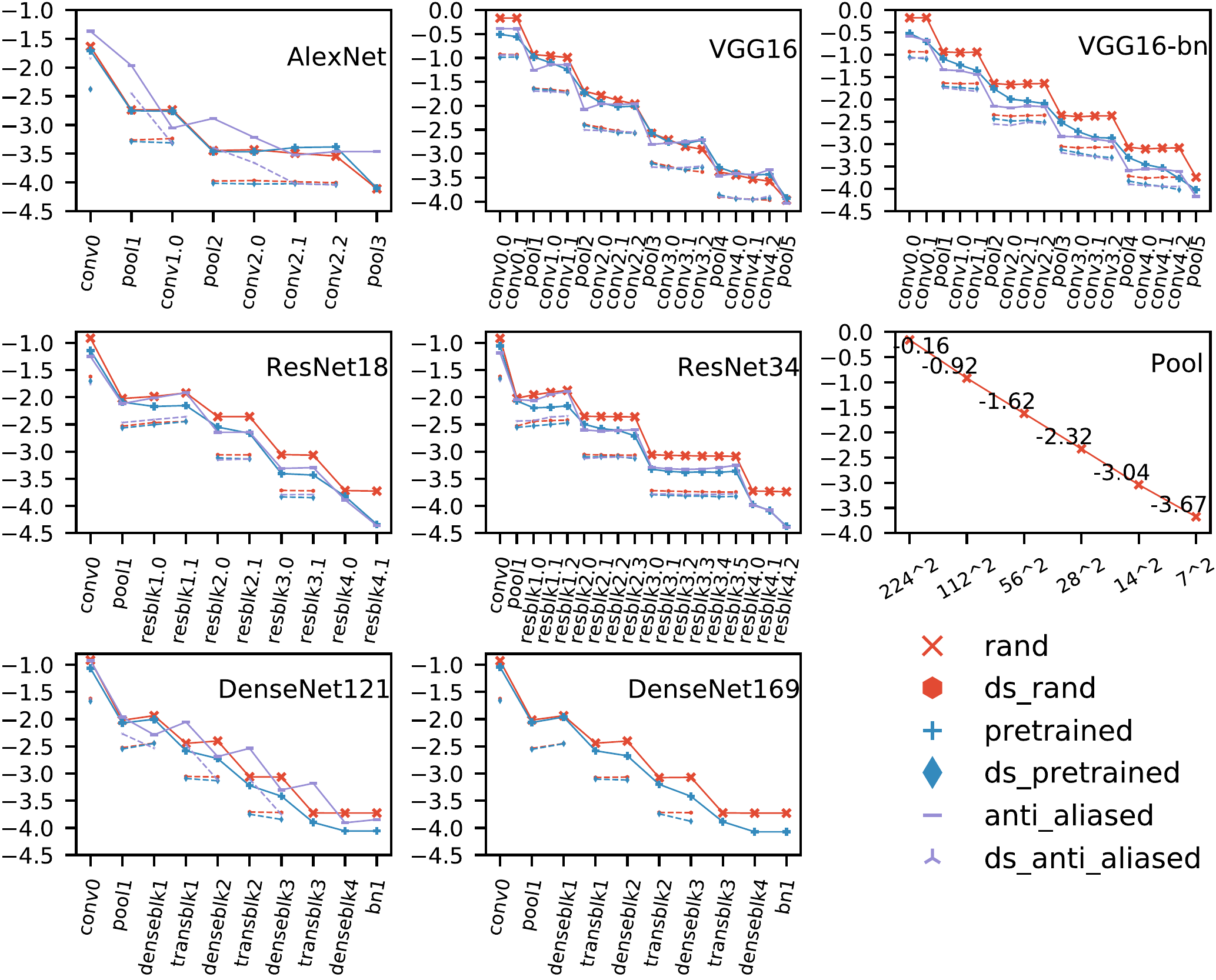}
    \end{center}
    \caption{SROP mean values in randomized and pre-trained backbones. The y axis denotes $log (SROP)$. Anti-aliased models are pre-trained. The benchmark \textit{Pool} is a max-pooling layer whose kernel size and stride are two. SROP curves with the \textit{ds} prefix are computed by replacing intermediate and downscaling layers with the benchmark \textit{Pool}.}
    \label{fig:flow}
\end{figure*}

Figure \ref{fig:base} demonstrates the SROP variations caused by downscaling blocks and a BN layer. Figure \ref{fig:flow} visualized untrained and pre-trained SROP profiles of various models and the simulation results of replacing intermediate layers by benchmark pooling. We can report the SROPs in the useful information variation caused by model architectures. Under the SROP framework, a decline in SROP represents an increment of useful information. Sufficient training increases the useful information in data representations, which is verified by the reduction of SROPs.

\paragraph{With and without downscaling blocks/a BN layer}
SROPs from max-pooling layers with large kernel size (3) edge close to the baseline ($kernel \quad size=2$) at a small output size in AlexNet. This confirms that large kernel size produces low SROPs in randomized blocks.

The only difference between VGG16 and VGG16-bn blocks is a BN layer. Outputs of layers with a BN block have small SROP inter-quartile ranges in outputs, indicating that BN makes the layer-wise output more concentrated compared with the condition without BN. From the view of SROPs, we also observe that BN reduces the internal covariate shift \cite{bn}.
\paragraph{SROP variations from anti-aliased/intermediate layers}
Different SROP profiles are observed when using anti-aliased and intermediate layers. It's known that anti-aliased layers output anti-aliased and useful feature maps. Compared with the origin models, variations of useful information and SROPs are both observed. A finding is that pre-trained anti-aliased layers can preserve more high-frequency components at initial layers instead of losing information albeit employing additional low-pass filters. In particular, the mean SROP values of \textit{resblk1.1} in randomized and pre-trained anti-aliased ResNet18 are almost the same. It proves that a downscaling layer can reduce the frequency of useful information in preceding feature maps. Anti-aliased models have low SROPs at the end of the hidden layer except for AlexNet, which is in accordance with their excellent performance.

Intermediate layers improve lower-frequency representations when the spatial resolution stays constant. This is reflected in lower SROPs in the downscaling layers when intermediate layers are provided. For instance, SROPs in \textit{resblk1.0} grow higher after applying \textit{resblk1.1}, but pre-trained intermediate layers enable lower SROPs in \textit{resblk2.0} compared with downscaling directly with the benchmark. It can be reported in a SROP scheme that intermediate layers improve the learnability of useful information. The increase of intermediate layers obtains smooth SROP profile, as shown by the SROP curves of ResNet18 and ResNet34.
\paragraph{SROP profiles and model architectures}
The diversity of SROP profiles reflects the diversity of designs for improving the extraction of useful information. Untrained and pre-trained models possess similar SROP patterns. This condition indicates that model architectures predetermine the flow of useful information. The skip connection in the ResNet family injects high-frequency components, embodied as an uptrend in the layers of the same spatial resolution. Anti-aliased blocks induce different SROP patterns. Retaining sufficient high-frequency information at shallow layers accounts for the gain in performance through training anti-aliased models.

\paragraph{SROPs in randomized and pre-trained models}
The SROP dip in deep layers is the main difference between the randomized and pre-trained models. The SROP discrepancy verifies that a well-trained model converts useful information to low-Nyquist-frequency representations.
\paragraph{Minimum mean SROPs and spatial resolution}
There exists a required low-end of sampling frequency to carry all useful information. LFP implies that the indispensable spatial resolution is low. Table \ref{tab:output_size_SROP} explores the minimum mean SROP at each output size across the architectures mentioned in Table \ref{tab:layer_name}. Modern pre-trained models can approximately cut SROP values in half compared with randomized models, especially when the output size is less than $56\times 56$. This condition indicates that low-frequency information will be perfectly preserved after successive downscaling blocks. The constraint of LFP encourages low-frequency representations before downscaling operations. It is true even for DenseNet and ResNet, where the downscaling layers can encode and preserve high-frequency components compared to maxpooling layers. Previous ideas emphasize the loss of high-frequency information caused by downscaling operations~\cite{LearningIT,make_conv_shift_inva}. However, our results show that the loss of useful information from downscaling steps is negligible. Therefore, given that all useful information is transmissible, additional filters are likely to introduce redundancy for large models. It may explain the diminishing accuracy gain from an increased model width. The minimum mean SROP is higher than $50\%$ of the baseline when output sizes are larger than $56\times 56$, as shown in Table \ref{tab:output_size_SROP}. This finding indicates that transforming useful information to low-frequency representations is difficult at initial layers.
\begin{table*}
    \begin{center}
    \begin{tabular}{l c c c c}
    \hline
    \textbf{Output size} &\textbf{Baseline}& \textbf{Min mean SROP} & \textbf{Layer} & \textbf{Anti-aliased}\\
    \hline
      $224 \times 224$&0.852 &0.494 & conv0.1(VGG16-bn)&No \\
      $112 \times 112$ &0.399&0.237 & conv1.1 (VGG16-bn) &Yes \\
      $56 \times 56$ ($55 \times 55$)&0.198 &0.101&denseblk1 (DenseNet121)&Yes  \\
      $28 \times 28$ ($27 \times 27$)& 0.098 & 0.047  & conv1.0 (AlexNet)&Yes \\
      $14 \times 14$ ($13 \times 13$) &0.048&0.023& conv4.2 (VGG16-bn) &No  \\
      $7 \times 7$ &0.025& 0.012 &resblk4.2 (ResNet34) &Yes\\
      \hline
    \end{tabular}
    \end{center}
    \caption{The minimum mean SROP across model backbones pre-trained on ImageNet at different spatial resolutions. The output sizes of AlexNet feature maps are slight smaller, as displayed in brackets.}
    \label{tab:output_size_SROP}
\end{table*}
\subsection{Comparisons between SROPs and ID}\label{subsec:compar}
Layer-wise observation demonstrates the feasibility of SROPs when estimating useful information. Ansuini \etal reported hunch-backed layer-wise ID profiles. The largest ID is in the relative depth range of 0.2-0.4. Its value is approximately four times the ID at the first layer. Details about ID observations may be found in~\cite{Intrinsic_dimension_of_data_representation}. Meanwhile, SROP profiles are concave with a decaying spatial resolution in some error range, and the largest SROP value appears at initial layers. SROPs contain less redundancy than ID in the estimation of useful information. We give up follow-up experiments since ID is incompatible with the property of useful information (Sec. \ref{sec:truth_ui}).

\section{Discussion}
\label{sec:discussion}

\paragraph{Feasibility of SROP statistics} The feasibility of using SROP mean values is queried in Sec. \ref{secsec:statistics} and validated in Sec. \ref{sec:results}. Other SROP statistics can characterise layer-wise useful information from different perspectives. For instance, SROP variance or standard deviation measures the credibility of kernel disequilibrium or internal covariate shift in a DNN layer. In summary, SROP statistics are informative to describe useful information.

\paragraph{Potential applications of SROPs}
Sec. \ref{sec:results_input} shows that different SROP distributions among samples indicate different useful patterns in the overall dataset. Figure \ref{fig:srop_acc} reveals the potential applications of SROPs in adversarial learning tasks, where recognizing useful information is important. Kernel smoothness in the first convolutions of DNN contributes to adversarial robustness \cite{High-Frequency-Component-Helps-Explain-the}.

Monitoring the training process is a potential use of SROP variance and standard deviation because a small internal covariate shift is a desired property for DNNs \cite{bn}. Kernel-wise SROPs provide more detailed information compared with layer-wise SROPs. The useful information can be monitored channel-wisely by SROP.

Another potential application of SROP distribution is to evaluate whether a customized dataset is feasible for deep learning tasks. Obtaining the same SROP distribution of data representations as the benchmark database is a practical target for customized datasets. For instance, the peak with the highest SROPs in Figure \ref{fig:frog} when $w=0$ is composed of 893 images of category "1" and tens of other digit samples when using the MNIST dataset. Therefore, images of category "1" should be collected if the high-SROP peak is not observed in the customized dataset.

\paragraph{Limitations of SROPs}
The use of SROPs is subject to some limitations. Correctness of SROP variations remains unconfirmed in semi-trained and superficially trained models, where useful information remains vague and accuracy is an undependable estimation (e.g. some models fail to learn useful information in medical images \cite{med_image}). Strategies to decrease SROPs are supposed to improve model performance in these CASEs. Another limitation is that SROP is only relevant for topologically ordered layers like convolutional layers, because spectrum analysis is only meaningful when neighbour pixels have spatial correlations. Distinguishing low-frequency disturbance and information is a question raised in Sec. \ref{subsec:ui_var_input}. Expanding features of layer-wise SROP distributions is a possible solution to distinguish useful and irrelevant information.

\section{Conclusion}\label{sec:conclusion}
The use of the variations of useful information is proposed to explore useful information, and SROPs is propounded as a novel estimation based on the low-Nyquist-frequency nature of useful information. To extend the implementation of SROPs from a 1-D signal to 3-D feature maps, we utilize the prior knowledge that well-trained models should possess rotation invariance and small internal covariate shift. Sanity checks prove that SROP variations can accurately assess useful information variations, since layer-wise SROPs and useful information vary synchronously when model input and structures become different or model becomes well-trained. SROP is more rigorous because it contains less redundant information compared to ID, which indicates that the estimation of useful information by SROPs is more accurate. SROPs can be used to estimate channel-wise useful information, which has potential applications in various scenarios. It is a promising and convenient tool to explain why DNNs can achieve excellent performance.
\section*{Acknowledgements}  
This work was supported by National Natural Science Foundation of China (81473579 and 81973744), Beijing Natural Science Foundation (7173267).

{\small
\bibliographystyle{icml2020}
\bibliography{srop}
}





\end{document}